\RequirePackage{snapshot}
\documentclass[10pt,twocolumn,letterpaper]{article}
\usepackage{cvpr}
\usepackage{times}
\usepackage{epsfig}
\usepackage{graphicx}
\usepackage{amsmath}
\usepackage{amssymb}
\usepackage{lipsum}
\usepackage{xspace}
\usepackage{multirow}
\usepackage{pbox}
\usepackage{diagbox}
\usepackage{subcaption}
\usepackage{adjustbox}
\usepackage{colortbl}
\usepackage{afterpage}
\usepackage[%
pagebackref=true,breaklinks=true,%
letterpaper=true,colorlinks,bookmarks=false]{hyperref}
\usepackage{cleveref}

\newcommand{\ba}{\mathbf{a}}
\newcommand{\bb}{\mathbf{b}}

\newcommand{\bx}{\mathbf{x}}

\newcommand{\bz}{\mathbf{z}}

\def\methodname{ANet\xspace}
\newcommand{\myparagraph}[1]{\vspace{0.15cm}\noindent\textbf{#1.}}

\cvprfinalcopy % *** Uncomment this line for the final submission

\ifcvprfinal\fi

\title{AnchorNet: A Weakly Supervised Network \\ to Learn Geometry-sensitive Features For Semantic Matching}

\author{
David Novotny$^{1,2}$ ~ ~ Diane Larlus$^2$ ~ ~ Andrea Vedaldi$^1$ \\
\begin{minipage}{.5\textwidth}
\centering
$^1$\small{Visual Geometry Group\\Dept. of Engineering Science, University of Oxford\\}
{\tt\small \{david,vedaldi\}@robots.ox.ac.uk}
\end{minipage} 
\begin{minipage}{.5\textwidth}
\centering
$^2$\small{Computer Vision Group\\Xerox Research Centre Europe\\} 
{\tt\small diane.larlus@xrce.xerox.com}
\end{minipage}
}

\begin{document}
% --------------------------------------------------------------------
\maketitle
\begin{abstract} 
Despite significant progress of deep learning in recent years, 
state-of-the-art semantic matching methods still rely on legacy features such as SIFT or HoG. We argue that the strong invariance properties that are key to the success of recent deep architectures on the classification task make them unfit for dense correspondence tasks, unless a large amount of supervision is used. In this work, we propose a deep network, termed AnchorNet, that produces image representations that are well-suited for semantic matching. It relies on a set of filters whose response is geometrically consistent across different object instances, even in the presence of strong intra-class, scale, or viewpoint variations. Trained only with weak image-level labels, the final representation successfully captures information about the object structure and improves results of state-of-the-art semantic matching methods such as the deformable spatial pyramid or the proposal flow methods.
We show positive results on the cross-instance matching task where different instances of the same object category are matched as well as
on a new cross-category semantic matching task aligning pairs of instances each from a different object class.
\end{abstract}

\section{Introduction}\label{sec:intro}

Matching, i.e.\ the problem of establishing correspondences between images, is one of the tent-poles of image understanding. It is well known that, given matches between images of the same object or scene, it is possible to estimate 3D geometry (stereo and structure from motion) and motion (visual odometry, optical flow, and tracking).
But matching can be applied to much more abstract levels of understanding as well. For example, aligning different object instances of the same type~\cite{kim2013deformable,ham2016} allows to discover analogies between objects, inducting abstractions such as object categories.

\begin{figure}[t!]
  \centering
  \includegraphics[angle=90,origin=c,width=0.9\linewidth]{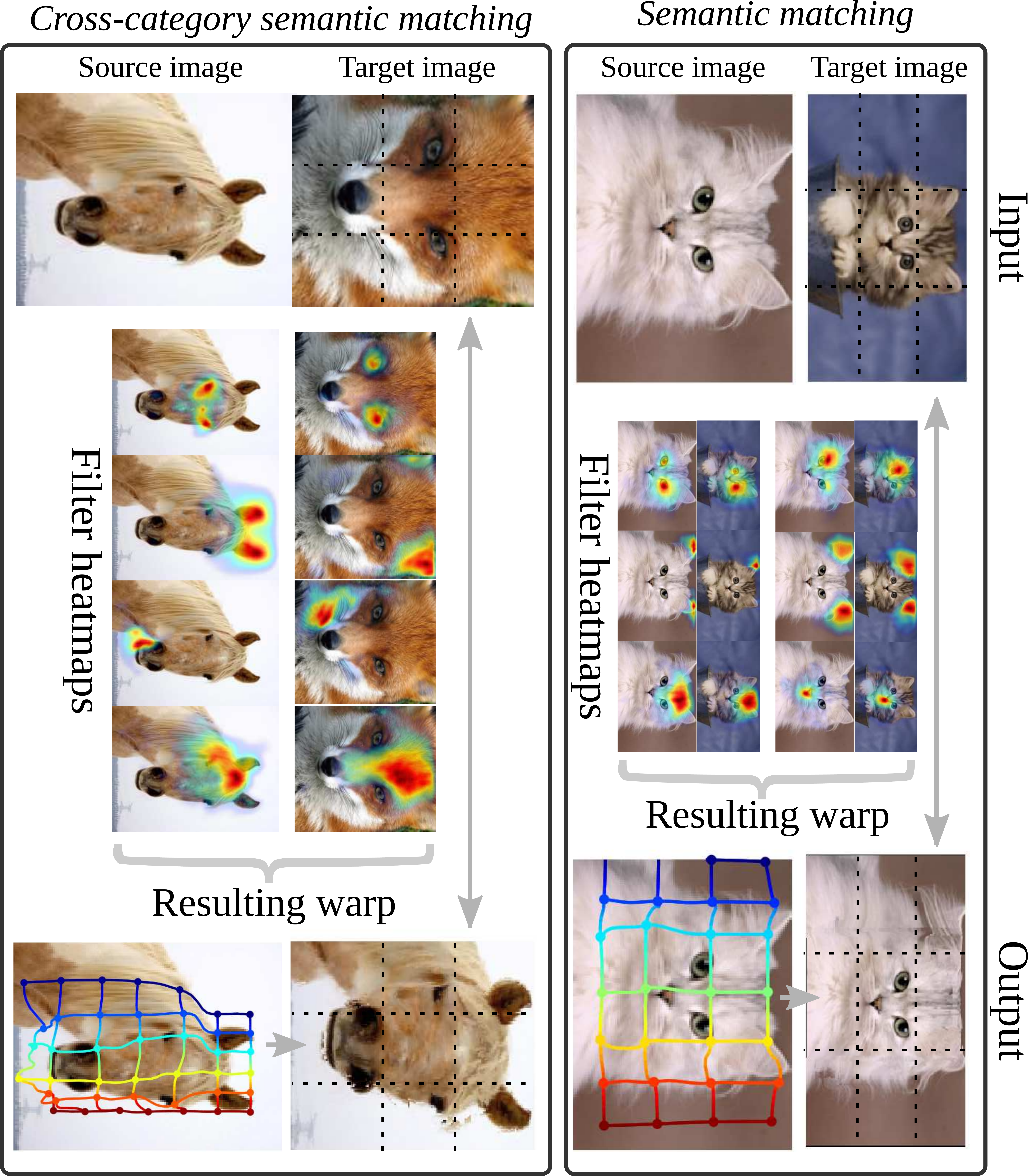}
  \caption{We propose AnchorNet, a novel deep architecture that produces an image representation which significantly improves state-of-the-art semantic matching methods. Key to its success is a set of filters with a sparse response that is geometrically consistent across different instances of a category or of two similar categories. Although these filters are learned in a weakly supervised manner (\ie only image-level labels are used) they tend to anchor reliably on meaningful object parts. }\label{f:splash}
  % \vspace{-0.24cm}
\end{figure}

While reliable techniques exist for low-level matching, high-level matching of different object instances remains a heavily-researched topic. Most of the work in this area has focused on finding powerful geometric regularizers, such as hierarchical correspondences~\cite{liu08siftflow} or deformable spatial pyramids~\cite{kim2013deformable}, to compensate for the still brittle visual descriptors. Surprisingly, even powerful convolutional neural network (CNN) descriptors have been found lacking for cross-instance matching~\cite{long2014do,ham2016,zhou2016learning}, and in fact comparable or even inferior to old hand-crafted features such as SIFT~\cite{lowe2004sift} and HoG~\cite{dalal05hog} for this task.

\begin{figure*}[t!]
	\newcommand{\filterimH}{4.05em}
	\newcommand{\agfilterW}{0.95\columnwidth}
 \scriptsize
  \begin{center}

  % \scalebox{0.9}{ 
  % {\renewcommand{\arraystretch}{-10}%
   \begin{tabular}{ccc}  
    \pbox[b]{\textwidth}{
	\includegraphics[height=\filterimH]{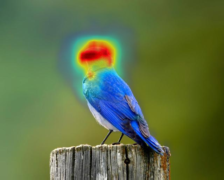}
	\includegraphics[height=\filterimH]{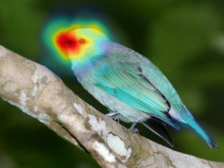}
	\includegraphics[height=\filterimH]{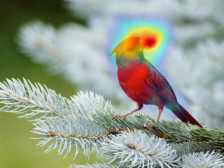}\\
	\includegraphics[height=\filterimH]{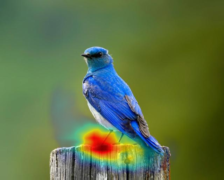}
	\includegraphics[height=\filterimH]{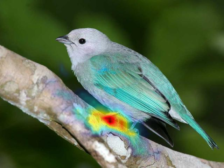}
	\includegraphics[height=\filterimH]{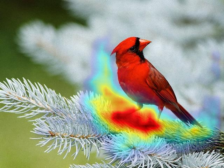}\\
	\includegraphics[height=\filterimH]{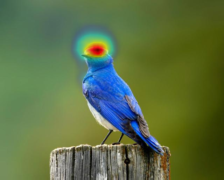}
	\includegraphics[height=\filterimH]{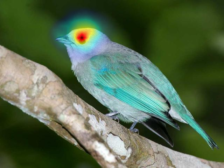}
	\includegraphics[height=\filterimH]{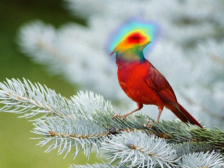}\\
	\includegraphics[height=\filterimH]{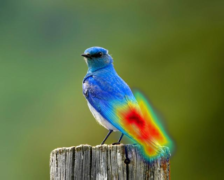}
	\includegraphics[height=\filterimH]{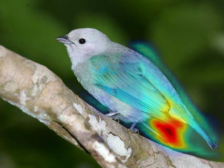}
	\includegraphics[height=\filterimH]{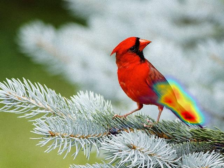}
    }
    &
    \pbox[b]{\textwidth}{
    \includegraphics[height=\filterimH]{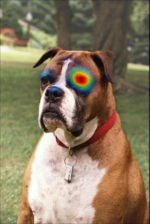}
	\includegraphics[height=\filterimH]{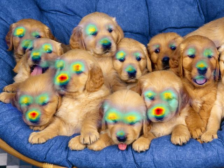}
	\includegraphics[height=\filterimH]{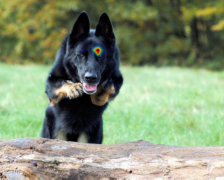}\\
	\includegraphics[height=\filterimH]{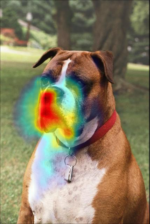}
	\includegraphics[height=\filterimH]{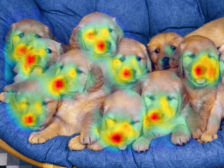}
	\includegraphics[height=\filterimH]{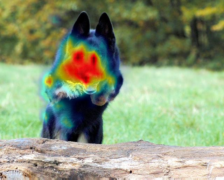}\\
	\includegraphics[height=\filterimH]{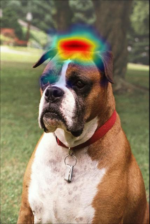}
	\includegraphics[height=\filterimH]{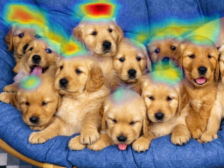}
	\includegraphics[height=\filterimH]{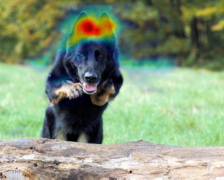}\\
	\includegraphics[height=\filterimH]{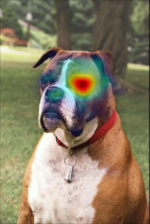}
	\includegraphics[height=\filterimH]{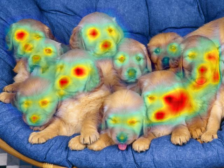}
	\includegraphics[height=\filterimH]{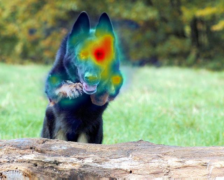}
    }
 %    &
 %    \pbox[b]{\textwidth}{
	% \includegraphics[height=\filterimH]{figures/filters/car/car_kp0013_im0001.png}
	% \includegraphics[height=\filterimH]{figures/filters/car/car_kp0013_im0002.png}
	% % \includegraphics[height=\filterimH]{figures/filters/car/car_kp0013_im0003.png}
	% \includegraphics[height=\filterimH]{figures/filters/car/car_kp0013_im0004.png}\\
	% \includegraphics[height=\filterimH]{figures/filters/car/car_kp0017_im0001.png}
	% \includegraphics[height=\filterimH]{figures/filters/car/car_kp0017_im0002.png}
	% % \includegraphics[height=\filterimH]{figures/filters/car/car_kp0017_im0003.png}
	% \includegraphics[height=\filterimH]{figures/filters/car/car_kp0017_im0004.png}\\
	% \includegraphics[height=\filterimH]{figures/filters/car/car_kp0031_im0001.png}
	% \includegraphics[height=\filterimH]{figures/filters/car/car_kp0031_im0002.png}
	% % \includegraphics[height=\filterimH]{figures/filters/car/car_kp0031_im0003.png}
	% \includegraphics[height=\filterimH]{figures/filters/car/car_kp0031_im0004.png}\\
	% \includegraphics[height=\filterimH]{figures/filters/car/car_kp0036_im0001.png}
	% \includegraphics[height=\filterimH]{figures/filters/car/car_kp0036_im0002.png}
	% % \includegraphics[height=\filterimH]{figures/filters/car/car_kp0036_im0003.png}
	% \includegraphics[height=\filterimH]{figures/filters/car/car_kp0036_im0004.png}
 %    }
    & 
	% \adjustbox{max width=12.3em} {
	\pbox[b]{\textwidth}{
		\adjustbox{max width=\agfilterW} { \pbox[b]{\columnwidth} {
			\includegraphics[height=0.83cm]{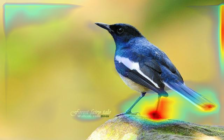}
			\includegraphics[height=0.83cm]{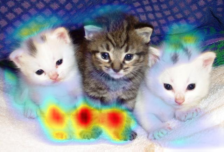}
			\includegraphics[height=0.83cm]{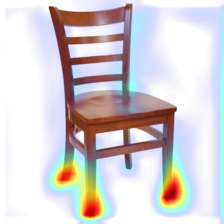}
			\includegraphics[height=0.83cm]{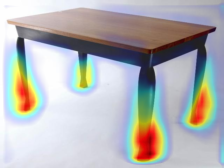}
			\includegraphics[height=0.83cm]{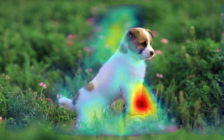}
			\includegraphics[height=0.83cm]{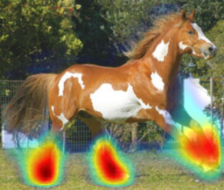}
			\includegraphics[height=0.83cm]{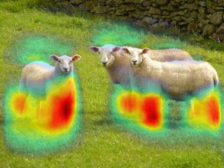}}} \\
		% \adjustbox{max width=\columnwidth}{ \pbox[b]{\columnwidth} {
	 % 		\includegraphics[height=1.1cm]{figures/filters/classag/kp0008_im0078.png}
		% 	\includegraphics[height=1.1cm]{figures/filters/classag/kp0008_im0096.png}
		% 	\includegraphics[height=1.1cm]{figures/filters/classag/kp0008_im0114.png}
		% 	\includegraphics[height=1.1cm]{figures/filters/classag/kp0008_im0144.png}
		% 	\includegraphics[height=1.1cm]{figures/filters/classag/kp0008_im0170.png} } } \\
		\adjustbox{max width=\agfilterW}{ \pbox[b]{\columnwidth} {
			\includegraphics[height=1.1cm]{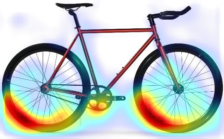}
			\includegraphics[height=1.1cm]{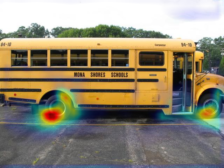}
			\includegraphics[height=1.1cm]{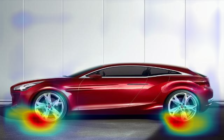}
			\includegraphics[height=1.1cm]{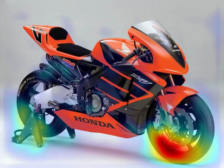}
			\includegraphics[height=1.1cm]{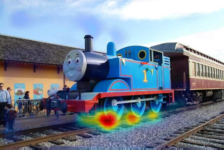}}} \\
		\adjustbox{max width=\agfilterW}{ \pbox[b]{\columnwidth} {
			\includegraphics[height=0.96cm]{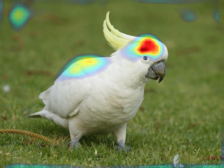}
			\includegraphics[height=0.96cm]{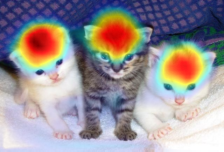}
			\includegraphics[height=0.96cm]{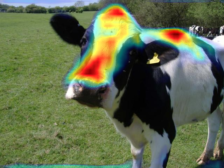}
			\includegraphics[height=0.96cm]{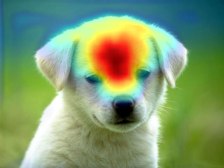}
			\includegraphics[height=0.96cm]{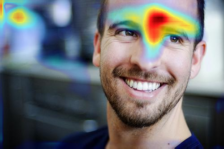}
			\includegraphics[height=0.96cm]{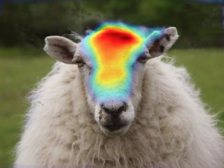}}} \\
		\adjustbox{max width=\agfilterW}{ \pbox[b]{\columnwidth} {
			\includegraphics[height=1.32cm]{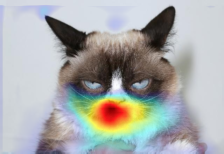}
			\includegraphics[height=1.32cm]{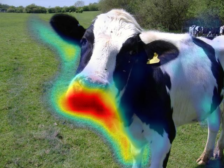}
			\includegraphics[height=1.32cm]{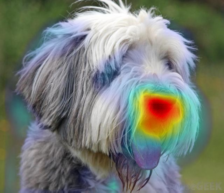}
			\includegraphics[height=1.32cm]{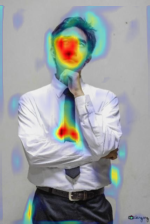}
			\includegraphics[height=1.32cm]{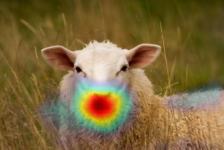}}}
		% \adjustbox{max width=\columnwidth}{ \pbox[b]{\columnwidth} {
		% 	% \includegraphics[height=1cm]{figures/filters/classag/kp0143_im0026.png}
		% 	\includegraphics[height=1cm]{figures/filters/classag/kp0143_im0076.png}
		% 	\includegraphics[height=1cm]{figures/filters/classag/kp0143_im0092.png}
		% 	\includegraphics[height=1cm]{figures/filters/classag/kp0143_im0112.png}
		% 	\includegraphics[height=1cm]{figures/filters/classag/kp0143_im0121.png}
		% 	\includegraphics[height=1cm]{figures/filters/classag/kp0143_im0166.png}}}
	 } \\
    {\normalsize(a)} & {\normalsize(b)} & {\normalsize(c)} \\ % & {\normalsize(d)} \\
   \end{tabular}
   % }
   \vspace{-0.6cm}
  \end{center}
  % \caption{Example anchor filters discovered by the AnchorNet. Subfigures (a) and (b) depict the responses of the class specific filters $F_k^{C_i}$ for the bird and car classes respectivelly (one filter per row); (c) contains the responses of the class agnostic filters  $F_k^{S}$ in images of different categories.}
  \caption{Example responses of anchor filters discovered by the AnchorNet. (a), (b) show the class specific filters $F_k^{C_i}$ for bird and dog classes respectively while (c) depicts the class agnostic filters $F_k^{S}$ across different categories (one filter per row).} 
  \label{fig:learnedfilters}
  \vspace{-0.2cm}
\end{figure*}

It is unclear why CNN representations, which perform well for many challenging vision tasks, including object detection~\cite{girshick15fastrcnn} and segmentation~\cite{long15fcn}, image captioning~\cite{vinyals15show}, and visual question answering~\cite{antol15VQA}, have not been found to work as well for cross-instance matching. Our hypothesis is that this is due to the fact that CNNs are trained on large datasets such as Imagenet ILSVRC~\cite{deng09imagenet}  purely for the image classification task. By learning with the sole purpose of predicting a global image label, CNNs become insensitive to local details and geometry and hence work poorly for matching. This effect can be reversed by fine-tuning the model on substantial amounts of data strongly supervised with bounding box~\cite{girshick15fastrcnn} or keypoint \cite{choy16universal} annotations. While this allows to use CNNs as excellent object and keypoint detectors, it defeats the purpose of using CNN features as generic descriptors for \emph{discovering} correspondences in an unsupervised manner, as matching requires.

In this paper, we address this issue by introducing a new deep architecture %and pre-training strategy 
that can learn \textit{representations that work well for cross-instance matching} (\Cref{f:splash}), while using \textit{exactly the same supervision} as traditional pre-training -- namely image-level labels used to train categorizers on ILSVRC12~\cite{deng09imagenet}. Using only image-level labels for matching amounts to weak supervision since the labels do not provide any information on the geometry of objects or scenes. 

Our key insight is that a set of \emph{diverse} and \emph{sparse} filter responses provides a powerful representation
for establishing matches. Convolutional features that respond sparsely on an image tend to automatically \emph{anchor} to distinctive image structures such as semantic object parts. Further enforcing diversity of the filter bank responses results in a good coverage. This yields a unique
description for \emph{all} object fragments which is an essential property that enables reliable estimation of \emph{dense} semantic correspondences.

We incorporate this idea by extracting from information-rich residual hypercolumns (\cref{s:hc}) a bank of distinctive and diverse filters with orthogonal responses (\cref{s:class-spec};~\Cref{fig:learnedfilters}). In this framework, which we call \emph{AnchorNet}, geometric consistency is not imposed explicitly, but emerges spontaneously. We also show how to compress banks of class-specific filters into a class-agnostic bank (\cref{s:agnostic}) which works well for all classes. %, including unseen ones.

Extensive experiments show that the proposed representation can be seamlessly leveraged by state-of-the-art semantic matching methods such as the Deformable Spatial Pyramid \cite{kim2013deformable} or Proposal Flow \cite{ham2016} in order to improve their performance (\cref{sec:expmatching}). For the first time, we also show that high-level correspondences can be established between objects of different categories, including new ones, unseen during the training of our network (\cref{sec:expDA}).

%!TEX root = paperCR.tex
\section{Related Work}
\label{sec:related}

\myparagraph{Finding dense correspondences}
The classical matching methods estimate very accurate pixel correspondences between two images of the same scene, in presence of moderate viewpoint variations \cite{horn93determining, matas02robust, okutomi1993}. Early methods use different hand-crafted features such as SIFT \cite{lowe2004sift}, HoG \cite{dalal05hog}, SURF \cite{bay2008surf} or DAISY \cite{Tola10daisy}. This task has many applications including stereo matching \cite{okutomi1993}, optical flow \cite{horn93determining,weinzaepfel13deepflow}, or wide baseline matching \cite{matas02robust,yang2014daisyfilterflow}.

Recent works have generalized the notion of flow to image pairs that are only semantically related \cite{liu11siftflow,qui14scale,kim2013deformable,tau15dense,ham2016}. This requires handling a higher degree of variability in appearance. 
The semantic alignment task also finds many applications such as image completion \cite{barnes10patchmatch}, enhancement \cite{hacohen2011non}, or segmentation \cite{liu11siftflow}, and video depth estimation \cite{karsch2012depth}.
The SIFT Flow algorithm \cite{liu08siftflow,liu11siftflow} pioneered the idea of dense correspondences across different scenes and proposes a multi-resolution image pyramid and a hierarchical optimization algorithm for efficiency. This approach got extended by the Deformable Spatial Pyramid (DSP) algorithm \cite{kim2013deformable} that introduced a multi-scale regularization with a hierarchically connected pyramid of graphs. The generalized deformable spatial pyramid \cite{hur15generalized} improves over DSP by enforcing additional spatial constraints at a significant computational cost. 
The Patch Match method \cite{barnes09patchmatch} and its extension \cite{barnes10patchmatch} target general purpose matching, including cross-instance matching.
The method of \cite{bristow2015dense} builds an exemplar-LDA classifier for every pixel to obtain dense correspondences that improve the performance of scene flows.
Proposal Flow \cite{ham2016} leverages the recent development in object proposals and uses local and geometric consistency constraints to establish dense semantic correspondences.
Finally, WarpNet \cite{kanazawa16warpnet} learns correspondences by exploiting the relationships within a fine-grained dataset. 

A few methods \cite{huang07unsupervised,huang12learning,peng2010rasl,kemelmacher2012collection,mobahi14a,zhou15flowweb} have posed the problem of finding correspondences as the joint alignment of multiple pairs of images, defining the task of collective alignment. These methods assume sets of 
images that share a category label and consistent viewpoints. 
The latest method in this field is FlowWeb \cite{zhou15flowweb}, that builds a fully connected graph with images as nodes, and pairwise flow fields as edges. Yet, this method scales poorly with the size of the image collection, and
it is not straightforward to establish pairwise alignments between new samples.

\myparagraph{Deep features for correspondences}
Long \etal \cite{long2014do} studied the application of CNN features pre-trained on large classification datasets for finding correspondences between object instances. 
They found that CNN features perform on par with hand-crafted alternatives such as SIFT for the weakly-supervised keypoint transfer problems, and can outperform them when keypoint supervision is available. 
This work paved the way to new deep architectures trained for finding dense correspondences between same object or scene instances \cite{dosovitskiy14flownet,zbontar16stereo,thewlis16fully-trainable}. 
Recently, Choy \etal \cite{choy16universal} proposed a deep architecture that performs well at cross-instance alignment, but requires 
strong supervision in form of many keypoint matches.

The question of training deep features without keypoint annotations still remains unanswered, as state-of-the art semantic matching methods \cite{kim2013deformable,ham2016} still rely on hand-engineered SIFT and HoG respectively.
\section{Method}\label{sec:method}

\begin{figure*}[ht!]
  \centering
  \includegraphics[width=1\linewidth]{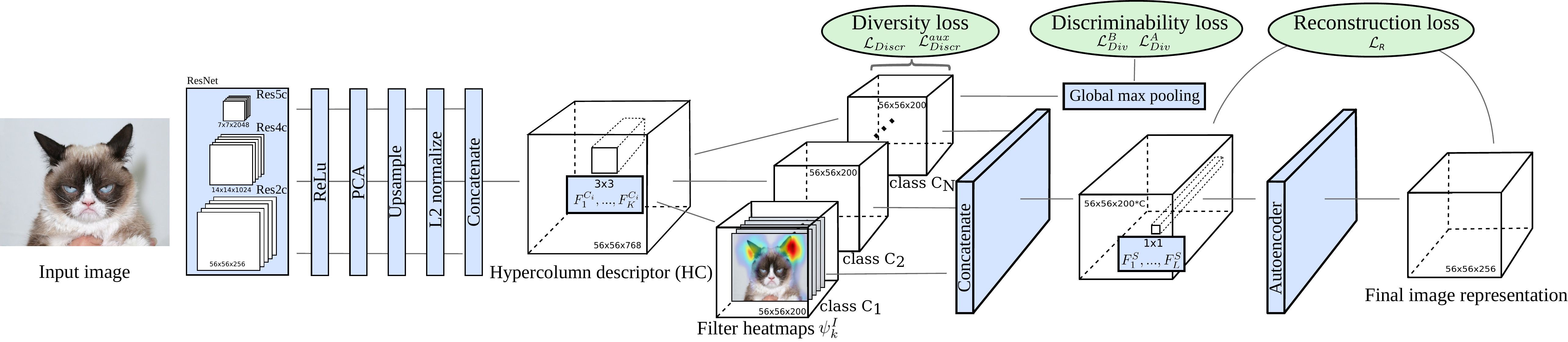}
  \caption{The proposed AnchorNet architecture. First, images are described using hypercolumn descriptors. Sparse filters are discovered for each category using a set of discriminability and diversity losses. Finally a denoising auto-encoder learns how to share these filters between categories, leading to a final category-agnostic representation generalizing to new classes.}
  \label{fig:architecture}
  \vspace{-0.2cm}
\end{figure*}

The output of a deep convolutional layer in a CNN is a tensor $\bx \in \mathbb{R}^{H\times W\times D}$ of height $H$, width $W$, and with $D$ feature channels. Thus, at each spatial location $(u,v)$, one obtains a $D$-dimensional feature vector $\mathbf{d}_{uv} = (x_{uv1}, \dots, x_{uvD})$. As noted by~\cite{cimpoi15deep}, such CNN feature vectors are analogous to hand-crafted dense descriptors like HoG and Dense-SIFT and can often be used as a plug-and-play replacement for the latter in applications. However, as noted in e.g. \cite{long2014do} and shown in the experiments, this substitution does not work well for cross-instance matching algorithms such as DSP \cite{kim2013deformable} and Proposal Flow \cite{ham2016}.

Since CNNs can be turned in excellent keypoint detectors by fine-tuning on data strongly annotated with keypoint labels~\cite{choy16universal,tulsiani2015viewpoints}, the reason for this failure must be in the way most CNNs are pre-trained on image classification tasks. Note that collecting keypoint annotations for every category does not scale and defeats the purpose of cross-instance matching, which is to discover such correspondences automatically. As a solution, we propose a new architecture that, while using the same image-level supervision as the standard pre-training on the classification task, learns features with better geometric awareness.

Our method is motivated by a simple observation. Suppose that learning encourages a feature to respond very locally (ideally a point). 
A convolutional filter can do this only by responding to a visual structure that occurs uniquely in each image -- hence the distinctive part or keypoint of an object. We call the latter the \emph{anchoring principle}. A geometry-aware representation suitable for semantic matching should discover such a complete set of features that ultimately covers the whole object. We can do so by learning a bank of filters that respond to complementary image locations. We call this the \emph{diversity principle}. Note that diversity indirectly encourages anchoring, as, if features respond to different parts of an image, they must also respond locally.
Armed with these insights, we propose next an architecture termed AnchorNet that follows the two principles. We then show that these are sufficient to significantly boost the geometric awareness of the resulting features. A diagram of our network is presented in \Cref{fig:architecture}.

\subsection{Residual hypercolumns}\label{s:hc}

We base our AnchorNet architecture on the powerful residual architectures of~\cite{he2015deep}. We select the ResNet50 model as a good compromise between speed and accuracy.

In order to improve the geometric sensitivity of the representation, we follow~\cite{hariharan15hypercolumns} and extract hypercolumns (HC). A HC $\mathbf{d}_{uv}$ at location $(u,v)$ in the image is created by concatenating the convolutional feature responses at that location for different layers of the network. 
Recall that, in most CNN architectures, deeper features have reduced resolution; HC compensates for this by upsampling the responses to a common size before concatenation. We denote the resulting network $\mathbf{d}=\Phi(I)$, where $I$ is the input image.

In more detail, we bilinearly upsample and concatenate the rectified outputs of the res2c, res4c and res5c layers~\cite{he2015deep} into a $56 \times 56 \times D$ hypercolumn tensor. Before concatenation, descriptors extracted at each layer are compressed by PCA to 256 dimensions (PCA is implemented as a $1\times 1$ filter bank) and $\ell^2$ normalized to balance their energies. This results in $D=768$ dimensional HC vectors. % Upsampling uses bilinear interpolation. 

\subsection{Learning anchoring features for an object type}\label{s:class-spec}

The residual HC are high-capacity descriptors reflecting both high-level semantics as well as low-level image details. While this suggests
that they should contain enough information for establishing matches,
their direct utilization leads to suboptimal results. Thus, we train a set of $3\times3$ convolutional filters $F_1, ..., F_K$ that compress the HC responses into a compact set of \emph{anchor filters} that are suitable for matching. To this end, we learn filters that satisfy two properties: discriminability and diversity.

\myparagraph{Discriminability constraints}
We start by learning filters $F_k$ predictive of an object category.
As a result, the filters tend to focus on relevant foreground objects, and rarely on the background.
Without loss of generality, we first consider a binary setting where images $I$ are either 
containing object instances of a single object category ($y_I = 1$) or irrelevant background ($y_I =-1$).
We later extend to multiple categories in \cref{s:agnostic}. 

Learning uses a large dataset of images with cheap-to-obtain image-level class labels. We follow common deep networks~\cite{simonyan2014very,he2015deep,krizhevsky12imagenet} and use ILSVRC~12~\cite{ILSVRC15} for training. % which makes our features directly comparable to theirs.
 
Discriminability is encouraged by minimizing the following loss function:
\begin{equation}
  \mathcal{L}_{\text{Discr}}(I,y_I;\Phi,F) 
  = - y_I \sum_{k=1}^K \operatorname{gmax} ~ \psi(F_k * \Phi(I)),
  \label{eq:discr1}
\end{equation}
where $\Phi(I)$ denotes the HC tensor extracted from image $I$. 
The function $\psi(z) = \log(1+\exp(z))$ is the smooth version of ReLU \cite{nair2010rectified}
and $\text{gmax}$ is the global max-pooling operator.

Minimizing $\mathcal{L}_{\text{Discr}}$ identifies the strongest response of each filter $F_k$ in the image and then enhances or suppresses it depending on whether the image contains the object. A disadvantage is that, due to the global max-pooling, the backpropagated signal is extremely sparse, which makes learning slow. To speed-up the convergence rate, we introduce a secondary loss function that, for negative images only, generates much denser gradients by using global average pooling ($\text{gavg}$) instead of max pooling:
\begin{equation}
  \mathcal{L}_{\text{Discr}}^{\text{aux}}(I,y_I;\Phi,F) = \delta_{[y_I = -1]} \sum_{k=1}^K \operatorname{gavg}  \max\{0,F_k * \Phi(I)\}
  \label{eq:discr2}.
\end{equation}
Using global average pooling is meaningful for the negative images, where all responses should be suppressed, but not for the positive ones, where only selected responses should be enhanced.

\myparagraph{Diversity constraints} Discriminability alone encourages filters to respond to the object; however different filters may learn to respond to redundant highly-distinctive object parts. In order to obtain good coverage (and ultimately good anchoring), we require the filters $F_k$ of one class to be active on \emph{diverse} regions.

The diversity constraint is implemented by two \emph{diversity losses} $\mathcal{L}_{\text{Div}}^A$ and $\mathcal{L}_{\text{Div}}^B$, encouraging orthogonality of the filters and of their responses, respectively. $\mathcal{L}_{Div}^A$ makes filters orthogonal by penalizing their correlations, as follows:
% \todo{either we fix the math as in VER1 and dont rerun anything or rerun exps and put there VER 2 (which is bit nicer)}
% {\color{green}
  % VERSION 1:
  \begin{equation}
    \mathcal{L}_{\text{Div}}^A(F) = \sum_{i \neq j} 
    \left|
    \sum_p
    \frac{\langle F_i^p, F_j^p \rangle}{\|F_i^p\|_F ~ \|F_j^p\|_F } 
    \right|
    \label{eq:div1}
  \end{equation}
where $F_i^p$ is the column of filter $F_i$ at spatial location $p$\footnote{\ie for our $3\times3$ filters $F_i$, $p \in \{1, 2, \dots, 9\}$}.
% }
% {\color{blue} \\
%   VERSION 2:
%   \begin{equation}
%     \mathcal{L}_{\text{Div}}^A(F) = \sum_{i \neq j}
%       \left|
%       \frac{\langle F_i, F_j \rangle}{\|F_i\|_F ~ \|F_j\|_F } 
%       \right|.
%     \label{eq:div1}
%   \end{equation}
% }
Note that orthogonal filters are likely to respond to different image structures, but this is not necessarily the case. Thus, we introduce a second term $\mathcal{L}_{\text{Div}}^B$ that directly 
decorrelates the filters' \emph{response maps} $\psi_k^I \doteq \psi(F_k * \Phi(I))$:
\begin{equation}
  \mathcal{L}_{\text{Div}}^B (I;\Phi,F) = \sum_{i \neq j}
  \left\lVert
  \frac{\langle \psi_i^I,  \psi_j^I\rangle} {\|\psi_i^I\|_F \|\psi_j^I\|_F}
  \right\rVert^2.
  \label{eq:div2}
\end{equation}
This term is further regularized by smoothing the response maps $\psi^I_k \doteq g_\sigma * \psi(F_k * \Phi(I))$ prior to computing the loss $\mathcal{L}_{\text{Div}}^B$, where $g_\sigma$ is a Gaussian kernel; this encourages filter responses to spread farther apart by dilating their activations.
% before these are pushed apart.
%
Note that inducing diversity among classifier prediction has been explored before \cite{gane2014learning,guzman2014efficiently,guzman2012multiple,schiegg2016learning,cai2011orthogonal},
however none of these works consider diversity as a loss to train a deep representation as we propose.

\myparagraph{Discussion} By making a large number of filters $F_k$ both discriminative and diverse, our method indirectly encourages them to become highly-specialized and hence to respond to unique parts of objects (the anchoring principle). This happens automatically, without enforcing such geometric properties explicitly.
This intuition is strongly supported by our experiments. Examples of the filters learned for the bird and dog classes are presented in \Cref{fig:learnedfilters} (a) and (b). It is apparent that filters fire on consistent object parts despite large intraclass variations, demonstrating the power of our formulation and its applicability to matching. 

\subsection{Class-agnostic representation}\label{s:agnostic}

In the previous section we have defined category specific anchoring filters.  In this section, we extend them to be generic to any category.
This allows to use the same representation for every image, irrespective of its label,
to match instances across different categories (\eg dog vs cat), and to even handle new categories.

First, a filter bank $F_1^{C_i}, ..., F_K^{C_i}$ is learned for each object category $C_1,\dots,C_N$ using the method above. Each object is learned by considering only images $C_i$ of that object class and a common background class $B$. Since filters are not learned to discriminate between objects, and since the diversity losses are applied only \emph{within} each bank, different filter banks can develop correlations. \Cref{fig:learnedfilters} illustrates this by showing that filters learned for the ``dog'' and ``bird'' classes capture similar concepts such as eyes or nose.

We take advantage of the overlap between different banks by introducing a new bank of $1\times1$ filters $F_1^S, ..., F_L^S$ that projects the class-specific responses of the filters $F_1^{C_1}, \dots, F_K^{C_N}$ to $L$ general-purpose response maps applicable to objects of any class.

In order to learn the projections $F^S$ end-to-end, we add a \emph{denoising autoencoder} (DAE) \cite{vincent2008extracting} to our architecture. DAE minimizes the~\emph{reconstruction loss} 
$\mathcal{L}_R (F^S, \hat\Gamma)$
\begin{equation}\label{e:enc}
\mathcal{L}_R (F^S, \hat\Gamma) = \mathcal{D}( \hat\Gamma , (F^S)^{\top} *  F^S * c(\hat\Gamma) )
\end{equation}
where  $\mathcal{D}(\ba,\bb) = \left\| {\ba}/{\|\ba\|}  - {\bb}/{\|\bb\|} \right\|^2$ is the $\ell^2$ distance between the $\ell^2$ normalized tensors $\ba$ and $\bb$ and $(F^S)^\top$ is the \emph{convolution transpose} operator~\cite{vedaldi15matconvnet}. %
Here $\hat \Gamma = \Gamma - \mu(\Gamma) $ denotes the stack of class-specific heatmaps $\Gamma = \operatorname{stack} ( \psi_{F_1^{C_1} }, \dots, \psi_{F_K^{C_N} } ) \in R^{W\times H\times (KN)}$ centered by removing their mean $\mu(\Gamma)$, estimated online during training. We have observed that centering followed by $\ell^2$ normalization greatly improves the convergence properties of $\mathcal{L}_R$. Function $c(\bz)$ injects noise by randomly setting to zero 25\% of the feature channels of the tensor $\bz$.

The decorrelation loss eq.~(\ref{eq:div1}) is  applied to the compression filters $F^S$ as well in order to encourage their diversity.

Note that the reconstruction loss $\mathcal{L}_R$, when optimized end-to-end with the rest of the model, encourages the maps $\hat \Gamma$ to shrink (because, if $\hat \Gamma=0$ everywhere, then the autoencoder has a trivial optimum). This is however prevented by the decorrelation losses $\mathcal{L}_{\text{Div}}^A$, $\mathcal{L}_{\text{Div}}^B$. $\mathcal{L}_R$ thus works as a regularizer enforcing part sharing. Examples of the learned class agnostic filters are in \cref{fig:learnedfilters} (c).

Denoising autoencoders have been used for domain adaptation before \cite{chen2012marginalized,glorot2011domain}. In a similar spirit, the last part of our network transforms a set of class (domain) specific filters into a domain invariant representation that can accommodate for any class, even the one not seen during training. %This can be seen as a domain generalization task.

\myparagraph{Network training}
AnchorNet is optimized with stochastic gradient descent (SGD) by minimizing
the sum of the proposed losses
$\mathcal{L}_{\text{Discr}}$, $\mathcal{L}_{\text{Discr}}^{\text{aux}}$, $\mathcal{L}_{\text{Div}}^A$, $\mathcal{L}_{\text{Div}}^B$ and $\mathcal{L}_{R}$,
with mini batches of size 16, a learning rate of $10^{-2}$, and a momentum of 0.0005.
Parameters of the network are initialized with the ResNet50 model pre-trained on ILSVRC12.
We use two-stage optimization to speed up the training process.
First, the class-specific filters $F_i^{C_k}$ are trained on $4\times10^4$ training images independently for each
object class $C_k$ keeping the rest of the network parameters fixed.
Then, we attach the autoencoder and the reconstruction loss to fine-tune all the network parameters end-to-end on $12\times10^3$ images.
Further details are provided in the supplementary material.

\newcommand{\tb}[1]{\textbf{#1}}
\definecolor{Gray}{gray}{0.95}
\newcolumntype{a}{>{\columncolor{Gray}}c}

\section{Experiments}\label{sec:exp}

\begin{table*}[t]
  \centering
  \setlength\tabcolsep{3pt}
  \scriptsize
\begin{tabular}{lacccccccccccccccccccc}
\hline
&  \textbf{mean} & aero & bike & bird & boat & bottle & bus  & car  & cat  & chair & cow  & dog  & horse & mbike & person & plant & sheep & sofa & table & train & tv \\ \hline
\multicolumn{22}{c}{Pairwise alignment methods}\\
\hline
% \rowcolor{green}
\textbf{DSP + \methodname-class}   & \tb{0.45}    &\tb{0.31} & \tb{0.49} & \tb{0.32} & 0.53 & \tb{0.75}   & \tb{0.51} & \tb{0.47} & 0.23 & 0.53 & \tb{0.37} & 0.20 & 0.33 & \tb{0.41} & \tb{0.22} & \tb{0.46} & \tb{0.45} & \tb{0.77} & 0.45 & 0.48 & \tb{0.74} \\
% \rowcolor{green}
\textbf{DSP + \methodname}         & \tb{0.45}   & 0.29 & 0.47 & 0.29 & 0.52 & 0.73   & 0.50 & 0.46 & \tb{0.25} & 0.53 & \tb{0.37} & \tb{0.21} & \tb{0.34} & 0.39 & 0.20 & 0.44 & 0.45 & \tb{0.77} & 0.45 & 0.51 & \tb{0.74}  \\
% \rowcolor{cyan}
% \textbf{DSP + \methodname-class-cleanNorm} & \tb{0.45} & \tb{0.31} & \tb{0.50} & \tb{0.31} & 0.51 & \tb{0.75} & \tb{0.51} & \tb{0.48} & 0.22 & 0.53 & \tb{0.37} & 0.19 & \tb{0.33} & \tb{0.41} & \tb{0.21} & \tb{0.45} & 0.43 & \tb{0.77} & 0.45 & 0.48 & \tb{0.74} \\
% \rowcolor{cyan}
% \textbf{DSP + \methodname-cleanNorm}      &  0.44 & 0.29 & 0.48 & 0.28 & \tb{0.52} & 0.73 & 0.49 & 0.46 & \tb{0.24} & 0.53 & 0.36 & \tb{0.20} & \tb{0.33} & 0.40 & \tb{0.21} & \tb{0.45} & \tb{0.44} & \tb{0.77} & 0.44 & \tb{0.50} & 0.73 \\
% DSP + resnet4               & 0.28 & 0.43 & 0.23 & 0.50 & 0.73   & 0.47 & 0.43 & 0.20 & 0.52  & 0.31 & 0.15 & 0.27  & 0.34  & 0.19   & 0.39  & 0.36  & 0.74 & 0.44  & 0.48  & 0.65 & 0.41 \\
% DSP + resnet5               & 0.27 & 0.42 & 0.23 & 0.50 & 0.73   & 0.47 & 0.42 & 0.20 & 0.51  & 0.31 & 0.15 & 0.26  & 0.33  & 0.19   & 0.39  & 0.35  & 0.74 & 0.44  & 0.48  & 0.65 & 0.40 \\
DSP + HC                          & 0.41      & 0.29 & 0.45 & 0.24 & 0.51 & 0.73 & 0.48 & 0.44   & 0.20 & 0.52 & 0.32 & 0.16 & 0.28 & 0.35 & 0.19 & 0.39 & 0.37 & 0.74 & 0.44 & 0.48 & 0.67\\ 
DSP + SIFT \cite{kim2013deformable}  & 0.39  & 0.25 & 0.46 & 0.21 & 0.48 & 0.63   & 0.50 & 0.45 & 0.19 & 0.48 & 0.30 & 0.14 & 0.26  & 0.35  & 0.13   & 0.40  & 0.37  & 0.66 & 0.37  & 0.48  & 0.62 \\ \hline
\textbf{Proposal Flow + \methodname-class}  & 0.43 & 0.26 & 0.43 & 0.28 & \tb{0.54} & 0.71   & 0.50 & 0.45 & 0.24 & \tb{0.54} & 0.32 & \tb{0.21} & 0.28  & 0.35  & 0.21   & 0.45  & 0.40  & 0.74 & \tb{0.46}  & 0.50  & 0.70 \\
\textbf{Proposal Flow + \methodname}  & 0.42    & 0.26  & 0.41 &  0.26 &  0.53 &  0.70 &  0.49 &  0.45 &  \tb{0.25} & \tb{0.54} & 0.31 &  0.19 &  0.28 &  0.31 &  0.17 &  0.43 &  0.39 &  0.74 &  0.44 &  \tb{0.52} & 0.69 \\
% Proposal Flow + Res4        & 0.27 & 0.44 & 0.26 & 0.54 & 0.70   & 0.50 & 0.45 & 0.23 & 0.53  & 0.32 & 0.18 & 0.28  & 0.33  & 0.17   & 0.44  & 0.39  & 0.74 & 0.45  & 0.52  & 0.66 & 0.42 \\
% Proposal Flow + Res5        & 0.23 & 0.34 & 0.25 & 0.53 & 0.70   & 0.47 & 0.43 & 0.22 & 0.52  & 0.30 & 0.18 & 0.26  & 0.27  & 0.17   & 0.41  & 0.38  & 0.73 & 0.45  & 0.49  & 0.60 & 0.39 \\
Proposal Flow + HC                  & 0.42  & 0.26 & 0.42 & 0.26 & \tb{0.54} & 0.70 & 0.50 & 0.45 & 0.23 & 0.53 & 0.32 & 0.18 & 0.27 & 0.32 & 0.18 & 0.43 & 0.38 & 0.74 & 0.45 & 0.51 & 0.64  \\
Proposal Flow + HoG \cite{ham2016} & 0.41    &0.25  & 0.45 & 0.23 & \tb{0.54} & 0.70 & 0.49 & 0.44   & 0.19 & 0.53 & 0.30 & 0.16 & 0.25 & 0.35 & 0.16 & 0.41 & 0.35 & 0.74 & 0.44 & 0.50 & 0.63 \\  \hline
Baseline: NoFlow &  0.39 &  0.27 &  0.40 &  0.22 &  0.50 &  0.73 &  0.46 &  0.42 &  0.20 &  0.51 &  0.30 &  0.15 &  0.25  & 0.32 &  0.18 &  0.38 &  0.34 &  0.74 &  0.44 &  0.47 &  0.64  \\ \hline
%\hline
\multicolumn{22}{c}{Collective alignment methods}\\
\hline
FlowWeb \cite{zhou15flowweb} & 0.43  & 0.33 & 0.53 & 0.24 & 0.51 & 0.72   & 0.54 & 0.51 & 0.20 & 0.52  & 0.32 & 0.15 & 0.29  & 0.45  & 0.19   & 0.41  & 0.39  & 0.73 & 0.41  & 0.51  & 0.68  \\ \hline
\end{tabular}
\caption{Weighted IoU for pairwise \textbf{semantic part matching} on PASCAL Parts. The proposed methods are in \textbf{bold}.}
\label{tab:segtransfer}
\end{table*}

We thoroughly compare our method with existing techniques for semantic matching (\cref{sec:expmatching}). Then, we assess how well our features allow to establish matches across images of different categories (\cref{sec:expDA}) which, to the best of our knowledge, was never demonstrated before.

Note that for all reported results, \emph{training only uses ILSVRC12}~\cite{deng09imagenet} images and labels, where the categories are merged according to the PASCAL-ILSVRC class mapping from \cite{deng09imagenet} (\eg \textit{sofa} is a merge of ``studio couch'' and ``day bed''). In this manner, 231 ILSVRC classes are used as positive examples spread over the 20 PASCAL VOC classes; the remaining 769 classes are used to form the set $B$ of negative (background) images. Even when we report results on one of the $N=20$ PASCAL VOC \cite{everingham10voc} classes, \textit{none} of the PASCAL VOC training data is used. 

\subsection{Dense pairwise semantic matching}\label{sec:expmatching}

We follow the standard practice~\cite{zhou15flowweb,ham2016} of using a dataset with manually annotated semantic keypoints or regions and assess how well a semantic matching method in combination with different types of features transfers the annotations from an image to another. We experiment on three datasets following their evaluation protocol.

\begin{figure*}[t]
\input{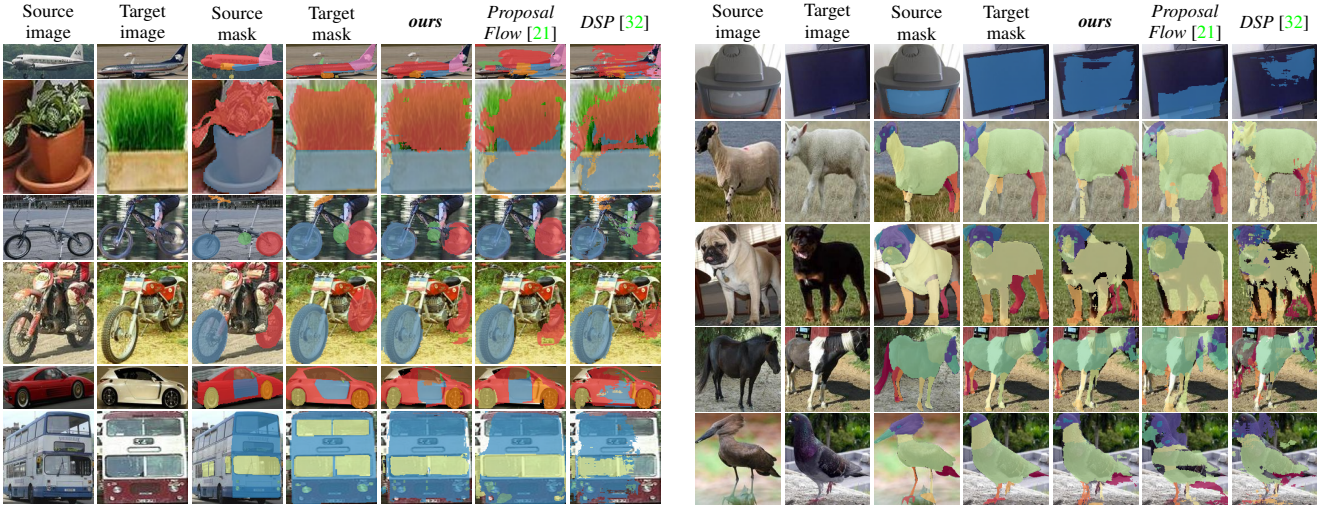}
 \caption{ \textbf{Segmentation mask transfer} on PASCAL Parts for DSP+\methodname~(ours), Proposal Flow + HoG, and DSP + SIFT.}
 \vspace{-0.3cm}
\label{fig:segtransfer}
\end{figure*}

\begin{table*}[t]
  \centering
  \setlength\tabcolsep{5pt}
  \scriptsize
  \begin{tabular}{lacccccccccccc} \hline
 & \textbf{mean} & aero & bike & boat & bottle & bus  & car  & chair & mbike & sofa & table & train & tv   \\
\hline
\multicolumn{14}{c}{Pairwise alignment methods}\\
\hline
% \rowcolor{green}
\textbf{DSP + \methodname-class}    & \tb{0.24}   &\tb{0.23} & 0.28 &\tb{ 0.06}& 0.38 & \tb{0.44} & \tb{0.39} & \tb{0.14} & \tb{0.19} & 0.16 & 0.11 & 0.13 & \tb{0.41}  \\
% \rowcolor{green}
\textbf{DSP + \methodname}          & 0.23 & 0.22 & 0.25 & 0.06 & 0.35 & 0.42 & 0.34 & \tb{0.14} & 0.17 & \tb{0.17} & \tb{0.13} & \tb{0.14} & 0.40 \\
% \rowcolor{cyan}
% \textbf{DSP + \methodname-class-cleanNorm} & \tb{0.25} & \tb{0.24} & 0.29 & 0.05 & \tb{0.40} & \tb{0.45} & \tb{0.40} & \tb{0.14} & \tb{0.19} & \tb{0.17} & \tb{0.12} & 0.13 & \tb{0.43}\\
% \rowcolor{cyan}
% \textbf{DSP + \methodname-v2-cleanNorm}    & 0.24 & 0.22 & 0.25 & \tb{0.06} & 0.36 & 0.42 & 0.34 & 0.14 & 0.18 & 0.17 & 0.14 & \tb{0.14} & 0.41\\
% }
DSP + HC                   & 0.20  & 0.20 & 0.23 & 0.05 & \tb{0.39} & 0.36 & 0.25 & 0.10 & 0.15 & 0.12 & 0.10 & 0.12 & 0.28 \\ 
DSP + SIFT     \cite{kim2013deformable}           & 0.18  & 0.17 & \tb{0.30} & 0.05 & 0.19 & 0.33 & 0.34 & 0.09 & 0.17 & 0.12 & 0.09 & 0.12 & 0.18  \\ \hline
\textbf{Proposal Flow + \methodname-class}  & 0.17 & 0.17 & 0.21 & 0.05 & 0.25 & 0.26 & 0.27 & 0.10 & 0.14 & 0.12 & 0.07 & 0.10 & 0.24 \\
\textbf{Proposal Flow + \methodname}    & 0.16   & 0.16 & 0.19 &  0.05 &  0.22 &  0.26 &  0.25 &  0.10 &  0.12  & 0.11 &  0.05 &  0.12 &  0.23 \\
Proposal Flow + HC       & 0.16    & 0.17 & 0.21 & 0.05 & 0.23 & 0.27 & 0.24 & 0.09 & 0.13 & 0.12 & 0.05 & 0.11 & 0.20    \\ 
Proposal Flow + HoG   \cite{ham2016} & 0.17 & 0.20 & 0.26 & 0.05 & 0.20 & 0.31 & 0.29 & 0.10 & 0.17 & 0.13 & 0.05 & 0.13 & 0.21  \\ \hline
Baseline: NoFlow  & 0.17 & 0.18 & 0.17 &  0.05 & \tb{0.39} & 0.31 & 0.17 &   0.09  & 0.12  & 0.11 & 0.07 & 0.11 & 0.24 \\ \hline
% DSP + resnet4               & 0.20 & 0.22 & 0.05 & 0.39 & 0.35 & 0.24 & 0.10 & 0.14 & 0.11 & 0.09 & 0.12 & 0.27 & 0.19 \\
% DSP + resnet5               & 0.19 & 0.19 & 0.05 & 0.38 & 0.32 & 0.19 & 0.09 & 0.13 & 0.11 & 0.08 & 0.11 & 0.25 & 0.17 \\
% Proposal Flow + Res4        & 0.19 & 0.24 & 0.05 & 0.23 & 0.28 & 0.27 & 0.09 & 0.15 & 0.12 & 0.05 & 0.13 & 0.21 & 0.17 \\
% Proposal Flow + Res5        & 0.13 & 0.11 & 0.04 & 0.21 & 0.21 & 0.19 & 0.07 & 0.08 & 0.08 & 0.05 & 0.09 & 0.14 & 0.11 \\
\hline
\multicolumn{14}{c}{Collective alignment methods}\\
\hline
FlowWeb \cite{zhou15flowweb} & 0.26  & 0.29 & 0.41 & 0.05 & 0.34   & 0.54 & 0.50 & 0.14  & 0.21  & 0.16 & 0.04  & 0.15  & 0.33  \\ \hline
  \end{tabular}
\caption{PCK ($\alpha = 0.05$) for semantic keypoint transfer on the 12 rigid classes of the PASCAL Parts dataset.}
\label{tab:kptransfer}
\end{table*}

\myparagraph{Compared methods} 
The most successful cross-instance matching methods include DSP \cite{kim2013deformable} and Proposal Flow \cite{ham2016} (PF). In their original formulation, these methods performed best with the Dense SIFT \cite{lowe2004sift} feature for DSP, and the whitened version of HoG \cite{hariharan2012discriminative} for PF. In the following experiments, we replace these descriptors with our representation, as follows.

For DSP, the learned filter banks produce a dense field of feature vectors which are bilinearly upsampled to the original image size, $\ell_2$ normalized and passed to DSP as a plug-and-play replacement of Dense SIFT.  For PF, we mimic their use of HoG: every object proposal serves as a pooling region for the set of filter activations that are extracted once for every image. The pooling is performed by reading-off the filter activations inside the region and resizing them to $8\times8$ using bilinear interpolation. This tensor is then vectorized and $\ell^2$ normalized to form the final descriptor of the proposal region.
We use the variant of PF that extracts 1000 selective search boxes \cite{uijlings2013selective} per image.
The rest of the matching procedure is identical to the original PF algorithm. 

We compare both the class-agnostic (\textbf{\methodname}) and class-specific (\textbf{\methodname-class}) variants of our anchor filters. 
The class-agnostic variant uses the 256 dimensional features produced by the autoencoder filters $F^S$, whereas
\methodname-class uses the output of the class-specific filters $F^{C_i}$ corresponding to a given PASCAL VOC object category $C_i$. Thus, \methodname-class assumes
knowledge of the object class label while \methodname is universally applicable without requiring additional image-specific information.
As baseline descriptors we consider SIFT, HoG and HC descriptors formed by concatenating the PCA projected layers of ResNet50 (res2c, res4c and res5c - \cref{s:hc}).
We also report the NoFlow baseline that predicts zero-displacement for every pixel. %It allows to measure the difficulty of different datasets and tasks.

While we focus on pairwise matching, an alternative is to align many images together, known as co-alignment. Among various co-alignment methods, including \cite{huang07unsupervised,peng2010rasl,kemelmacher2012collection}, FlowWeb~\cite{zhou15flowweb} is currently the state of the art. Due to its superior performance, we only report results for FlowWeb; however, while FlowWeb works very well, it is important to note that it is also substantially more expensive than pairwise matching, does not scale well and cannot accommodate for new image pairs. 

\myparagraph{Evaluation of segmentation masks transfer} We compare the various methods on the task of transferring semantic part segmentation masks, strictly following the protocol of~\cite{zhou15flowweb}. Dense semantic matches, as determined by DSP or PF given a descriptor, are used to warp the part segmentation mask from a source to a target image. The matching quality is assessed as the average weighted intersection-over-union (IoU) between the predicted masks and the ground-truth ones for different semantic parts. 
% Note that mask transfer is a ``one-shot'' task, and is very different from \emph{object or part segmentation or detection}. 
The results are reported in~\Cref{tab:segtransfer}, qualitative results are provided in \Cref{fig:segtransfer}.

We make the following observations. 
First, the ResNet50 features, perform at most marginally better, than SIFT or HoG, while both \methodname and \methodname-class features
improve performance for both DSP (+6\% IoU) and PF (+1\% IoU). % to benefit from the deep features.
Second, the class-specific features~\methodname-class perform on par with the class-agnostic features \methodname, 
demonstrating the ability of our domain generalization approach to compress the class-specific filters into the class-agnostic ones.
Third, our features, in combination with DSP, exhibit the best average performance among all the compared methods. Remarkably, both \methodname~and \methodname-class outperform all co-alignment methods, including FlowWeb \cite{zhou15flowweb}, achieving state-of-the-art results on this dataset. This is an interesting finding as the co-alignment methods exploit the small viewpoint and appearance variations in order to improve pairwise alignments.

\myparagraph{Evaluation of keypoint matching}
We also evaluate performance on matching semantic keypoints. Corresponding annotations are provided by~\cite{xiang2014beyond} for the 12 rigid PASCAL VOC categories. Similar to the previous section, we use the dataset from \cite{zhou15flowweb}, and, strictly following their evaluation protocol, we assess the matching accuracy using PCK, setting the misalignment tolerance parameter $\alpha$ to $0.05$.

\Cref{tab:kptransfer} contains the results of this experiment. Our features improve the original DSP results by a large margin (+6\% PCK), obtaining state-of-the-art results on this dataset among the pairwise alignment methods. Pairwise matching becomes in fact competitive with the results obtained by FlowWeb in co-alignment, although the latter use more information. Proposal Flow is generally weaker on this task and is not helped by the better features. 

\myparagraph{Evaluation of region matching}
As a third benchmark dataset, we use the PF dataset and corresponding protocol as described in detail in \cite{ham2016}. 
The dataset contains 10 image sets of 4 object types and the task is to establish matches between annotated semantic regions within the image sets.
We report region matching precision using the definitions specified in~\cite{ham2016}. 
\Cref{tab:pf} contains the results obtained by using the code and data made available by \cite{ham2016}. 

We evaluate our deep features in combination with the two matching methods presented in \cite{ham2016}: the best performing local offset matching (LOM), and the naive appearance matching (NAM). \methodname ~ is compared with the best performing feature from \cite{ham2016}, \ie HoG \cite{hariharan2012discriminative}. We observe that using \methodname-class features %(both \methodname and \methodname-class) 
in combination with both matching methods (LOM, NAM) brings a significant performance improvement.  
Note in particular that \methodname-class is sufficiently powerful to make the NAM baseline, which does not use any sophisticated geometric reasoning, competitive with the LOM+HoG, which uses geometric reasoning but handcrafted features (LOM+\methodname-class is even better).

% \begin{table}
%   \centering
%   \setlength\tabcolsep{3pt}
%   \scriptsize
%   \begin{tabular}{l|ccc|ccc}
%     \hline
%     & \multicolumn{3}{c|}{AuCs for PCR}                      & \multicolumn{3}{c}{AuCs for mIoU@k} \\
%     \hline
%     \diagbox{Matching}{Feature}      & HOG  & \methodname  & \methodname-class  & HOG & \methodname & \methodname-class \\
%     \hline
%     NAM: baseline  & 0.29 & 0.36    & \textbf{0.41}      & 41.0 & 35.8 & \textbf{42.4} \\
%     LOM: Proposal Flow  & 0.43 & 0.43     & \textbf{0.46}      & 55.6 & 53.4   & \textbf{57.9} \\ \hline
% \end{tabular}
% \caption{On the Proposal Flow dataset.}
% \label{tab:pf}
% \end{table}

\begin{table}
  \centering
  \setlength\tabcolsep{3pt}
  \footnotesize
  \begin{tabular}{l|ccc}
    \hline
    & \multicolumn{3}{c}{AuCs for PCR} \\
    \hline
    \diagbox{Matching}{Feature}  & \textbf{\methodname-class} &  \textbf{\methodname}   & HOG \cite{ham2016} \\
    \hline
    NAM: baseline      & \textbf{0.41} & 0.36  & 0.29        \\
    LOM: Proposal Flow & \textbf{0.46} & 0.43  & 0.43     \\ \hline
\end{tabular}
\caption{\textbf{Region matching} on the PF dataset.
%Our proposed methods are in \textbf{bold}.
\vspace{-0.3cm}}
\label{tab:pf}
\end{table}

\begin{table*}[t]
  \centering
  \setlength\tabcolsep{5pt}
  \scriptsize
\begin{tabular}{laccccccccccccc} \hline
Source class           & &  bicycle   & mbike & bus & car & bus & dog & cat & sheep & dog   & horse & cow   & sheep & cow  \\ 
               &  \textbf{mean}     & $\downarrow$          & $\downarrow$          & $\downarrow$    & $\downarrow$    & $\downarrow$    & $\downarrow$    & $\downarrow$    & $\downarrow$      & $\downarrow$      & $\downarrow$      & $\downarrow$      & $\downarrow$      & $\downarrow$        \\
Target class           & &  mbike & bicycle   & car & bus & car & cat & dog & dog   & sheep & cow   & horse & cow   & sheep \\ \hline
\textbf{DSP + \methodname} & \tb{0.37} & \tb{0.35} & \tb{0.45} & \tb{0.52} & \tb{0.35} & 0.36 & \tb{0.25} & 0.25 & \tb{0.34} & \tb{0.27} & \tb{0.31} & \tb{0.47} & \tb{0.37} & \tb{0.58}  \\
DSP + HC                 &  0.32  & 0.27  &0.44&  0.48& 0.32& 0.34& 0.20& 0.21& 0.22  &0.23&  0.27  &0.40 &0.28 &0.54   \\ 
 DSP + SIFT \cite{kim2013deformable} & 0.29  & 0.28 & 0.40  & 0.40    & 0.27  & 0.30 & 0.16  & 0.16 & 0.20 & 0.19 & 0.26& 0.31  & 0.28 & 0.50 \\ \hline
\textbf{Proposal Flow + \methodname}& 0.35 & 0.32 & 0.38 & 0.50 & 0.32 & \tb{0.37} & 0.23 & \tb{0.27} & 0.30 & 0.25 & 0.29 & 0.41 & 0.32 & 0.53  \\ 
Proposal Flow + HC        & 0.33  & 0.31 & 0.34 & 0.49 & 0.29 & 0.35 & 0.22 & 0.24 & 0.28 & 0.23 & 0.28 & 0.41 & 0.32 & 0.53  \\ 
Proposal Flow + HOG \cite{ham2016}& 0.31 & 0.30 & 0.43 & 0.48 & 0.30 & 0.35 & 0.19 & 0.21 & 0.22 & 0.19 & 0.25 & 0.37 & 0.29 & 0.50  \\ \hline
Baseline: NoFlow           &  0.27 & 0.26 & 0.44 & 0.35 & 0.26 & 0.25 & 0.17 & 0.18 & 0.22 & 0.17 & 0.22 & 0.29 & 0.26 & 0.49 \\ \hline
% DSP + res5c             & 0.26 & 0.44 & 0.47 & 0.32 & 0.33 & 0.20 & 0.21 & 0.22 & 0.23 & 0.26 & 0.39 & 0.28 & 0.53 & 0.32 \\
% DSP + res4c           & 0.26 & 0.44 & 0.47 & 0.32 & 0.34 & 0.20 & 0.21 & 0.22 & 0.23 & 0.26 & 0.39 & 0.28 & 0.54 & 0.32 \\
% ProposalFlow + res5c    & 0.31 & 0.31 & 0.49 & 0.29 & 0.32 & 0.21 & 0.25 & 0.28 & 0.22 & 0.27 & 0.39 & 0.31 & 0.49 & 0.32 \\
% ProposalFlow + res4c    & 0.31 & 0.31 & 0.49 & 0.29 & 0.32 & 0.21 & 0.25 & 0.28 & 0.22 & 0.27 & 0.39 & 0.32 & 0.49 & 0.32 \\
\end{tabular}
\vspace{-0.2cm}
\caption{Weighted IoU for \textbf{cross instance semantic part matching} on PASCAL Parts.}% The methods that use our proposed features are in \textbf{bold}.}
\label{tab:transferpascalparts}
\vspace{-0.2cm}
\end{table*}

\begin{figure}
\centering
\scriptsize
\begin{tabular}{l|rrr|rrr|r} 
% \hline
% Matching Alg.         & \multicolumn{3}{c|}{Proposal Flow}   & \multicolumn{3}{c|}{DSP}                                        & NoFlow  \\ \hline
% Feature               & \methodname                          & HoG                      & HC      & \methodname & SIFT  & HC   & -       \\ \hline
% PCK ($\alpha$ = 0.05) & \textbf{0.13}                        & 0.06                     & 0.09    & 0.11        & 0.06  & 0.08 & 0.04    \\
% PCK ($\alpha$ = 0.1)  & \textbf{0.32}                        & 0.18                     & 0.25    & 0.24        & 0.12  & 0.18 & 0.12    \\ \hline
\hline
Matching Alg.         & \multicolumn{3}{c|}{DSP}             & \multicolumn{3}{c|}{Proposal Flow}                                        & NoFlow  \\ \hline
Feature               &    \textbf{\methodname}  & HC   & SIFT        & \textbf{\methodname}                        & HC      & HoG    & -       \\ \hline
PCK ($\alpha$ = 0.05) &    0.11         & 0.08 & 0.06        & \textbf{0.13}                      & 0.09    & 0.06   & 0.04    \\
PCK ($\alpha$ = 0.1)  &    0.24         & 0.18 & 0.12        & \textbf{0.32}                      & 0.25    & 0.18   & 0.12    \\ \hline
\end{tabular}
\captionof{table}{\textbf{Semantic matching} on the AnimalParts dataset. For each method, we report the average PCK over all possible 12x12 domain pairs. An overview of individual cross-category results can be found in \Cref{fig:animalmatrices}}
\label{tab:transferanimals}
\vspace{0.2cm}
\includegraphics[width=0.9\linewidth]{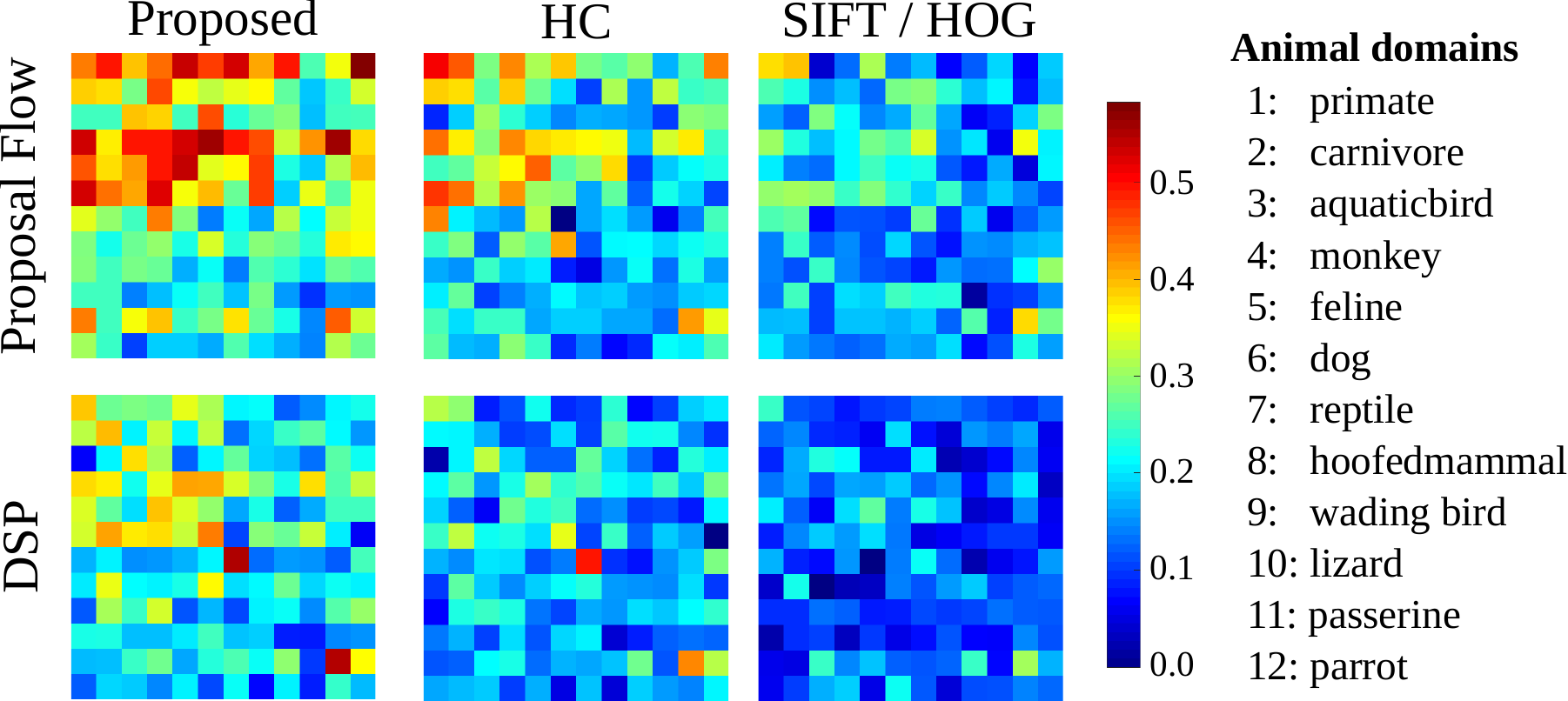}
\captionof{figure}{\textbf{Per-domain semantic matching} on the AnimalParts dataset. Cells are colored proportionally to the matching performance on a given animal class pair. Columns denote the source domains, 
rows the targets.}
\label{fig:animalmatrices}
\vspace{-0.35cm}
\end{figure} 

\begin{figure*}[t]
\input{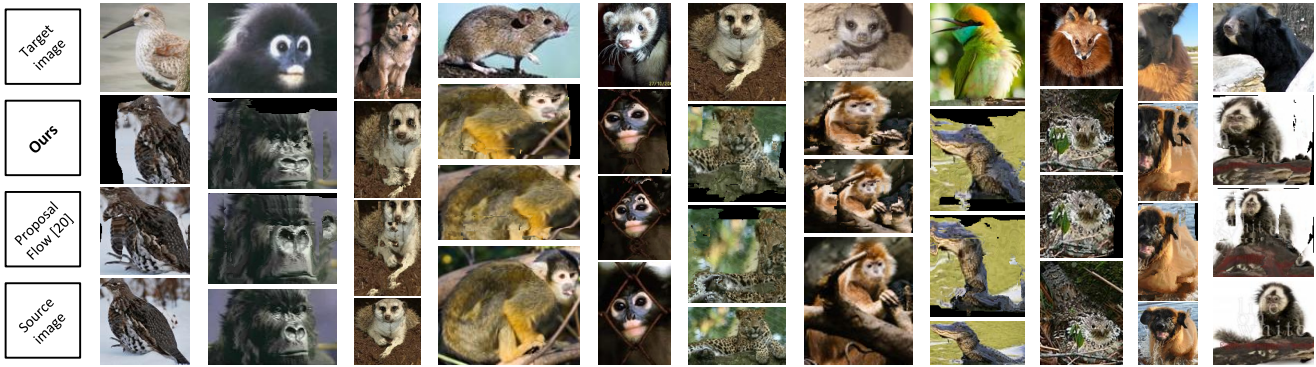}
 \caption{ \textbf{Cross-class alignments} on the AnimalParts dataset. Given a target (top row) and source images (bottom row)
we establish semantic correspondences between parts of animal classes. 
The alignment warps the source image into the target image.
We compare Proposal Flow + \methodname (ours - 2nd row) and Proposal Flow + HoG \cite{ham2016} (3rd row).
}
\label{fig:animalwarp}
\end{figure*}

\subsection{Generalization across categories}
\label{sec:expDA} 

The previous section experimented on the task of aligning different object instances of the same category.
Here, we depart from this scenario and consider instead \textit{cross-category matching},  where correspondences are established between objects of different categories. To the best of our knowledge, this is the first time this task is considered. 

For evaluation, we first use the PASCAL Parts \cite{chen2014detect} data from~\cite{zhou15flowweb}. Parts with different location qualifiers are merged into one (e.g.\ ``left-leg'' and ``right-leg'' are merged into ``leg'') to ensure shareability across categories. Overall, there are 9 object categories and 13 shared part types. 
% shareability is an existing word

% ON R3's request we put there some info about the animal parts dataset

% version 1
% Second, we consider the AnimalParts~\cite{novotny16i-have} dataset which includes only a few part types (``eye'' and ``foot''), but a large number of different categories -- 100 animals from the ILSVRC12 dataset. In order to present results compactly, animals are grouped in 12 families, based on the WordNet \cite{Miller1995} hierarchy. For each pair of super-classes, 40 image pairs are randomly sampled for evaluation, resulting in $\sim$7K image pairs in total. PCK is computed for each pair of super-classes, and the results are averaged over such pairs. The class-specific \methodname-class does not apply since the goal is to match across categories; instead, we compare \methodname to the other baselines.
% Note that the AnimalParts dataset was introduced in \cite{novotny16i-have} to study the transferability of semantic part detectors given a limited number of strong annotations. Here, we reuse the dataset in order to assess transferability of \methodname non-semantic filters trained without explicit supervision.

% version 2
Second, we consider the AnimalParts~\cite{novotny16i-have} dataset, introduced as a test-bed to study the transferability of semantic part detectors. % given a limited number of strong annotations.
Here, we reuse the dataset in order to assess transferability of \methodname filters trained without explicit supervision. 
AnimalParts includes only a few part types (``eye'' and ``foot''), but a large number of different categories -- 100 animals from the ILSVRC12 dataset. In order to present results compactly, animals are grouped in 12 families, based on the WordNet \cite{Miller1995} hierarchy. For each pair of super-classes, 40 image pairs are randomly sampled for evaluation, resulting in $\sim$7K image pairs in total. PCK is computed for each pair of super-classes, and the results are averaged over such pairs. The class-specific \methodname-class does not apply since the goal is to match across categories and most of these categories were not seen during training.
% ; instead, we compare \methodname to the other baselines.

\Cref{tab:transferpascalparts,tab:transferanimals} and \Cref{fig:animalmatrices} show that \methodname works substantially better than other matching methods. For the AnimalParts, the best results are obtained with Proposal Flow in combination with our features, with a 7\% PCK improvement over the PF + HoG baseline ($\alpha=0.05$). The fact that AnimalParts contains categories unseen at train time (\eg reptiles) demonstrates the scalability and generalization of the proposed approach.
For PASCAL Parts, similar to the intra-class matching experiment (\cref{sec:expmatching}), DSP performs best. Here~\methodname attains a 16\% relative improvement over the best previously published method (Proposal Flow + HoG). 
\Cref{fig:animalwarp} provides qualitative results. % on the AnimalParts dataset.

%!TEX root = paperCR.tex
\section{Conclusion}
\label{sec:ccl}
In this paper we have examined the problem of dense semantic matching.
Employing the concept of filter anchoring, we have designed a novel deep architecture, termed AnchorNet.
Supervised with only image-level labels, AnchorNet automatically learns a set of filters which respond in a sparse and geometrically consistent manner across object instances. Thanks to these filters, our architecture produces powerful representations for image matching. We experimentally validate these features in conjunction with state-of-the art semantic matching methods attaining state-of-the-art
performance on the segmentation transfer and keypoint matching tasks.
Versatility of our representation has been demonstrated on the new task of cross-category matching where we report positive results on two test-beds.

\myparagraph{Acknowledgments} We would like to thank Xerox Research Center Europe and ERC 677195-IDIU for supporting this research.

\appendix
\renewcommand{\thetable}{\Alph{table}}
\renewcommand{\thefigure}{\Alph{figure}}

\section{Learning details}

In this section we provide additional details about the learning protocol of AnchorNet.
Training converges after visiting $4 \times 10^4$ training samples (for each class) in stage 1 and $1.2 \times10^4$ samples in stage 2 (two days on a single GPU NVIDIA Tesla M40). 
The learning rate was fixed to a value of $10^{-2}$ with the minibatch size of 16 and the momentum set to the standard value of 0.0005.
The training data were augmented as in \cite{he2015deep}.

The losses were balanced as follows. 
The weights of $\mathcal{L}_{\text{Discr}}$ and $\mathcal{L}_{\text{Discr}}^\text{Aux}$ were set to $1$ and $10$ respectively.
The weight of $\mathcal{L}_R$ was set to a higher value of $10^6$ which is necessary due to the inhibition of the gradient by the $\ell_2$ normalization which takes place just before computing $\mathcal{L}_R$. The weights of $\mathcal{L}_{\text{Div}}^{A,B}$ and $\mathcal{L}_{\text{R}}$ were set to be as high as possible ($10^5$) such that 
$\mathcal{L}_{\text{Div}}^{A,B}\approx \mathcal{L}_{\text{R}} \approx 0$ are treated approximately as hard constraints. 
Importantly, $\mathcal{L}_R$ is optimized only during visiting positive samples as reconstructing the activations of negative samples would waste the capacity of the autoencoder.
During the first training stage, we sample positive and negative samples with equal probability. Furthermore, during stage 2, we ensure that the distribution of positive samples
is uniform over the set of 20 Pascal categories. This causes the positive samples from any given object category to be $20 \times$ less frequent than the negative samples. Hence, in order to rebalance losses in stage 2, we decrease the weights of negative samples by a factor of $20$.
Due to the fact that the gradients from $\mathcal{L}_{\text{Div}}^{A,B}$ exhibit high magnitudes, we decrease the learning rate on the layers bellow the first autoencoder layer
by a factor of $10^4$ during the second stage.

\section{Additional experimental results}

\Cref{tab:segtransfer,tab:kptransfer} provide an extension of Tables 1 and 2 from the paper. 
On top of the features already provided in Tables 1 and 2, we include more baseline features: res4c and res5c, which are extracted from the ResNet50 architecture
and the features from Simon et al. \cite{simon2015neural}. 
\cite{simon2015neural} selects part-like convolutional feature channels using a mixture of constellation models; however, if two different aspects are detected in two images, the set of common features is too sparse for matching. Thus, we converted their output to dense descriptors for use in DSP and PF by 1) modifying the HC from the ResNet50 architecture by retaining their part-like channels across all aspects (denoted as \textbf{Constellation-HC}) and 2) by backpropagating the part-like channel activations to the input image as they do, and using the image-level 
activations as dense descriptors (\textbf{Constellation-BP}).

Additionally, to quantify the impact of the diversity losses $\mathcal{L}_{\text{Div}}$, we also report the performance of the features produced by the \methodname-class method optimized without the diversity losses with DSP used as the matching algorithm (\textbf{DSP + \methodname-class w/o $\mathcal{L}_{\text{Div}}$}).

We observe that the res4c, res5c features as well as all the variants of the constellation features perform on par with the hypercolumn features (HC). 
The apparent drop in performance of DSP + \methodname-class w/o $\mathcal{L}_{\text{Div}}$ compared to DSP + \methodname-class highlights the contribution of the diversity losses.

% \section{Additional qualitative results}

% \myparagraph{Segmentation transfer on PascalParts}
% \Cref{fig:segtransfer} complements Figure 4 from the paper and contains additional qualitative results for the segmentation transfer task on the PASCAL Parts dataset.

% \myparagraph{Semantic matching on AnimalParts}
% \Cref{fig:animalwarp} complements Figure 6 from the paper and contains additional qualitative results for the semantic matching task on the AnimalParts dataset.

\begin{table*}[t]
  \centering
  \setlength\tabcolsep{3pt}
  \scriptsize
\begin{tabular}{lacccccccccccccccccccc}
\hline
&  \textbf{mean} & aero & bike & bird & boat & bottle & bus  & car  & cat  & chair & cow  & dog  & horse & mbike & person & plant & sheep & sofa & table & train & tv \\ \hline
\multicolumn{22}{c}{Pairwise alignment methods}\\
\hline
\textbf{DSP + \methodname-class}   & \tb{0.45}    &\tb{0.31} & \tb{0.49} & \tb{0.32} & 0.53 & \tb{0.75}   & \tb{0.51} & \tb{0.47} & 0.23 & 0.53 & \tb{0.37} & 0.20 & 0.33 & \tb{0.41} & \tb{0.22} & \tb{0.46} & \tb{0.45} & \tb{0.77} & 0.45 & 0.48 & \tb{0.74} \\
\textbf{DSP + \methodname-class w/o $\mathcal{L}_{\text{Div}}$} & 0.41 & 0.27 & 0.42 & 0.25 & 0.51 & 0.72 & 0.46 & 0.42 & 0.21 & 0.52 & 0.32 & 0.19 & 0.30 & 0.33 & 0.18 & 0.44 & 0.34 & 0.75 & 0.42 & 0.48 & 0.64  \\
\textbf{DSP + \methodname}         & \tb{0.45}   & 0.29 & 0.47 & 0.29 & 0.52 & 0.73   & 0.50 & 0.46 & \tb{0.25} & 0.53 & \tb{0.37} & \tb{0.21} & \tb{0.34} & 0.39 & 0.20 & 0.44 & 0.45 & \tb{0.77} & 0.45 & 0.51 & \tb{0.74}  \\
DSP + res4c          & 0.41     & 0.28 & 0.43 & 0.23 & 0.50 & 0.73   & 0.47 & 0.43 & 0.20 & 0.52  & 0.31 & 0.15 & 0.27  & 0.34  & 0.19   & 0.39  & 0.36  & 0.74 & 0.44  & 0.48  & 0.65  \\
DSP + res5c          & 0.40      & 0.27 & 0.42 & 0.23 & 0.50 & 0.73   & 0.47 & 0.42 & 0.20 & 0.51  & 0.31 & 0.15 & 0.26  & 0.33  & 0.19   & 0.39  & 0.35  & 0.74 & 0.44  & 0.48  & 0.65 \\
DSP + HC                          & 0.41      & 0.29 & 0.45 & 0.24 & 0.51 & 0.73 & 0.48 & 0.44   & 0.20 & 0.52 & 0.32 & 0.16 & 0.28 & 0.35 & 0.19 & 0.39 & 0.37 & 0.74 & 0.44 & 0.48 & 0.67\\ 
DSP + SIFT \cite{kim2013deformable}  & 0.39  & 0.25 & 0.46 & 0.21 & 0.48 & 0.63   & 0.50 & 0.45 & 0.19 & 0.48 & 0.30 & 0.14 & 0.26  & 0.35  & 0.13   & 0.40  & 0.37  & 0.66 & 0.37  & 0.48  & 0.62 \\ 
DSP + Constellation-HC & 0.40 & 0.28 & 0.42 & 0.23 & 0.50 & 0.73 & 0.47 & 0.42 & 0.20 & 0.52 & 0.31 & 0.15 & 0.27 & 0.34 & 0.19 & 0.38 & 0.36 & 0.74 & 0.44 & 0.48 & 0.65 \\
DSP + Constellation-BP & 0.40 & 0.27 & 0.41 & 0.23 & 0.50 & 0.73 & 0.46 & 0.42 & 0.20 & 0.51 & 0.31 & 0.15 & 0.26 & 0.33 & 0.18 & 0.38 & 0.35 & 0.73 & 0.44 & 0.47 & 0.64 \\
\hline
\textbf{Proposal Flow + \methodname-class}  & 0.43 & 0.26 & 0.43 & 0.28 & \tb{0.54} & 0.71   & 0.50 & 0.45 & 0.24 & \tb{0.54} & 0.32 & \tb{0.21} & 0.28  & 0.35  & 0.21   & 0.45  & 0.40  & 0.74 & \tb{0.46}  & 0.50  & 0.70 \\
\textbf{Proposal Flow + \methodname}  & 0.42    & 0.26  & 0.41 &  0.26 &  0.53 &  0.70 &  0.49 &  0.45 &  \tb{0.25} & \tb{0.54} & 0.31 &  0.19 &  0.28 &  0.31 &  0.17 &  0.43 &  0.39 &  0.74 &  0.44 &  \tb{0.52} & 0.69 \\
Proposal Flow + res4c     & 0.42   & 0.27 & 0.44 & 0.26 & 0.54 & 0.70   & 0.50 & 0.45 & 0.23 & 0.53  & 0.32 & 0.18 & 0.28  & 0.33  & 0.17   & 0.44  & 0.39  & 0.74 & 0.45  & \tb{0.52}  & 0.66  \\
Proposal Flow + res5c    & 0.39    & 0.23 & 0.34 & 0.25 & 0.53 & 0.70   & 0.47 & 0.43 & 0.22 & 0.52  & 0.30 & 0.18 & 0.26  & 0.27  & 0.17   & 0.41  & 0.38  & 0.73 & 0.45  & 0.49  & 0.60  \\
Proposal Flow + HC                  & 0.42  & 0.26 & 0.42 & 0.26 & \tb{0.54} & 0.70 & 0.50 & 0.45 & 0.23 & 0.53 & 0.32 & 0.18 & 0.27 & 0.32 & 0.18 & 0.43 & 0.38 & 0.74 & 0.45 & 0.51 & 0.64  \\
Proposal Flow + HoG \cite{ham2016} & 0.41    &0.25  & 0.45 & 0.23 & \tb{0.54} & 0.70 & 0.49 & 0.44   & 0.19 & 0.53 & 0.30 & 0.16 & 0.25 & 0.35 & 0.16 & 0.41 & 0.35 & 0.74 & 0.44 & 0.50 & 0.63 \\  
Proposal Flow + Constellation-HC  & 0.40 & 0.26 & 0.39 & 0.25 & 0.53 & 0.68 & 0.48 & 0.43 & 0.21 & 0.52 & 0.30 & 0.17 & 0.26 & 0.31 & 0.15 & 0.42 & 0.37 & 0.72 & 0.44 & 0.50 & 0.62 \\
Proposal Flow + Constellation-BP  & 0.39 & 0.25 & 0.38 & 0.23 & 0.53 & 0.68 & 0.47 & 0.41 & 0.20 & 0.51 & 0.29 & 0.16 & 0.25 & 0.30 & 0.15 & 0.41 & 0.35 & 0.71 & 0.43 & 0.49 & 0.60 \\
\hline
Baseline: NoFlow &  0.39 &  0.27 &  0.40 &  0.22 &  0.50 &  0.73 &  0.46 &  0.42 &  0.20 &  0.51 &  0.30 &  0.15 &  0.25  & 0.32 &  0.18 &  0.38 &  0.34 &  0.74 &  0.44 &  0.47 &  0.64  \\ \hline
%\hline
\multicolumn{22}{c}{Collective alignment methods}\\
\hline
FlowWeb \cite{zhou15flowweb} & 0.43  & 0.33 & 0.53 & 0.24 & 0.51 & 0.72   & 0.54 & 0.51 & 0.20 & 0.52  & 0.32 & 0.15 & 0.29  & 0.45  & 0.19   & 0.41  & 0.39  & 0.73 & 0.41  & 0.51  & 0.68  \\ \hline
\end{tabular}
\caption{Weighted IoU for pairwise \textbf{semantic part matching} (not to be confused with object or part detection or segmentation) on PASCAL Parts. The methods that use our proposed features are in \textbf{bold}.}
\label{tab:segtransfer}
\end{table*}

\begin{table*}[t]
  \centering
  \setlength\tabcolsep{5pt}
  \scriptsize
  \begin{tabular}{lacccccccccccc} \hline
 & \textbf{mean} & aero & bike & boat & bottle & bus  & car  & chair & mbike & sofa & table & train & tv   \\
\hline
\multicolumn{14}{c}{Pairwise alignment methods}\\
\hline
\textbf{DSP + \methodname-class}        & \tb{0.24}   &\tb{0.23} & 0.28 &\tb{ 0.06}& 0.38 & \tb{0.44} & \tb{0.39} & \tb{0.14} & \tb{0.19} & 0.16 & 0.11 & 0.13 & \tb{0.41}  \\
\textbf{DSP + \methodname-class w/o  $\mathcal{L}_{\text{Div}}$}  & 0.17 & 0.19 & 0.18 & 0.06 & 0.31 & 0.31 & 0.18 & 0.10 & 0.13 & 0.12 & 0.08 & 0.12 & 0.24 \\
\textbf{DSP + \methodname}               & 0.23 & 0.22 & 0.25 & 0.06 & 0.35 & 0.42 & 0.34 & \tb{0.14} & 0.17 & \tb{0.17} & \tb{0.13} & \tb{0.14} & 0.40  \\
DSP + HC                   & 0.20  & 0.20 & 0.23 & 0.05 & \tb{0.39} & 0.36 & 0.25 & 0.10 & 0.15 & 0.12 & 0.10 & 0.12 & 0.28 \\ 
DSP + res4c             & 0.19  & 0.20 & 0.22 & 0.05 & 0.39 & 0.35 & 0.24 & 0.10 & 0.14 & 0.11 & 0.09 & 0.12 & 0.27  \\
DSP + res5c             & 0.17   & 0.19 & 0.19 & 0.05 & 0.38 & 0.32 & 0.19 & 0.09 & 0.13 & 0.11 & 0.08 & 0.11 & 0.25 \\
DSP + SIFT     \cite{kim2013deformable}           & 0.18  & 0.17 & \tb{0.30} & 0.05 & 0.19 & 0.33 & 0.34 & 0.09 & 0.17 & 0.12 & 0.09 & 0.12 & 0.18  \\ 
DSP + Constellation-HC \cite{simon2015neural} & 0.18 & 0.20 & 0.21 & 0.05 & 0.39   & 0.33 & 0.20 & 0.10  & 0.13  & 0.12 & 0.09  & 0.12  & 0.26 \\
DSP + Constellation-BP \cite{simon2015neural} & 0.17 & 0.19 & 0.19 & 0.05 & 0.39   & 0.32 & 0.19 & 0.10  & 0.12  & 0.12 & 0.08  & 0.12  & 0.25 \\ \hline
\textbf{Proposal Flow + \methodname-class}  & 0.17 & 0.17 & 0.21 & 0.05 & 0.25 & 0.26 & 0.27 & 0.10 & 0.14 & 0.12 & 0.07 & 0.10 & 0.24 \\
\textbf{Proposal Flow + \methodname}    & 0.16   & 0.16 & 0.19 &  0.05 &  0.22 &  0.26 &  0.25 &  0.10 &  0.12  & 0.11 &  0.05 &  0.12 &  0.23 \\ 
Proposal Flow + HC       & 0.16    & 0.17 & 0.21 & 0.05 & 0.23 & 0.27 & 0.24 & 0.09 & 0.13 & 0.12 & 0.05 & 0.11 & 0.20    \\ 
Proposal Flow + res4c      & 0.17  & 0.19 & 0.24 & 0.05 & 0.23 & 0.28 & 0.27 & 0.09 & 0.15 & 0.12 & 0.05 & 0.13 & 0.21  \\
Proposal Flow + res5c       & 0.11 & 0.13 & 0.11 & 0.04 & 0.21 & 0.21 & 0.19 & 0.07 & 0.08 & 0.08 & 0.05 & 0.09 & 0.14  \\
Proposal Flow + HoG   \cite{ham2016} & 0.17 & 0.20 & 0.26 & 0.05 & 0.20 & 0.31 & 0.29 & 0.10 & 0.17 & 0.13 & 0.05 & 0.13 & 0.21  \\ 
Proposal Flow + Constellation-HC \cite{simon2015neural} & 0.14 & 0.18 & 0.17 & 0.04 & 0.19   & 0.25 & 0.20 & 0.08  & 0.12  & 0.10 & 0.05  & 0.10  & 0.17 \\
Proposal Flow + Constellation-BP \cite{simon2015neural} & 0.13 & 0.16 & 0.15 & 0.04 & 0.19   & 0.25 & 0.18 & 0.07  & 0.10  & 0.10 & 0.06  & 0.10  & 0.17 \\ \hline
Baseline: NoFlow  & 0.17 & 0.18 & 0.17 &  0.05 & \tb{0.39} & 0.31 & 0.17 &   0.09  & 0.12  & 0.11 & 0.07 & 0.11 & 0.24 \\ \hline
% DSP + resnet4               & 0.20 & 0.22 & 0.05 & 0.39 & 0.35 & 0.24 & 0.10 & 0.14 & 0.11 & 0.09 & 0.12 & 0.27 & 0.19 \\
% DSP + resnet5               & 0.19 & 0.19 & 0.05 & 0.38 & 0.32 & 0.19 & 0.09 & 0.13 & 0.11 & 0.08 & 0.11 & 0.25 & 0.17 \\
% Proposal Flow + Res4        & 0.19 & 0.24 & 0.05 & 0.23 & 0.28 & 0.27 & 0.09 & 0.15 & 0.12 & 0.05 & 0.13 & 0.21 & 0.17 \\
% Proposal Flow + Res5        & 0.13 & 0.11 & 0.04 & 0.21 & 0.21 & 0.19 & 0.07 & 0.08 & 0.08 & 0.05 & 0.09 & 0.14 & 0.11 \\
%\hline
\multicolumn{14}{c}{Collective alignment methods}\\
\hline
FlowWeb \cite{zhou15flowweb} & 0.26  & 0.29 & 0.41 & 0.05 & 0.34   & 0.54 & 0.50 & 0.14  & 0.21  & 0.16 & 0.04  & 0.15  & 0.33  \\ \hline
  \end{tabular}
\caption{PCK ($\alpha = 0.05$) for semantic keypoint transfer on the 12 rigid classes of the PASCAL Parts dataset.}
\label{tab:kptransfer}
\end{table*}

% \begin{figure*}[t]
% \input{figures/fig_heatmaps_supp}
% \caption{Example anchor filters discovered by the AnchorNet for the bicycle, bottle, bus, car, cat, chair, cow, dining table, horse classes.}
% \label{fig:learnedfilters_supp}
% \end{figure*}

% \begin{figure*}[t]
% \input{figures/fig_heatmaps_supp_contd}
% \caption{Example anchor filters discovered by the AnchorNet for the motorbike, person, potted plant, sheep, sofa, tv/monitor, train, boat, aeroplane classes.}
% \label{fig:learnedfilters_supp_contd}
% \end{figure*} 

% \begin{figure*}[t]
% \input{figures/segtransfersupp/fig_segtransfer}
% \caption{\textbf{Segmentation mask transfer} on PASCAL Parts for DSP+\methodname~(ours), Proposal Flow + HoG, and DSP + SIFT.}
% \label{fig:segtransfer}\end{figure*}

% \begin{figure*}[t]
% \input{figures/animaltransfer/fig_animals_supp}
%  \caption{ \textbf{Cross-class alignments} on the AnimalParts dataset. Given a target (top row) and source images (bottom row)
% we establish semantic correspondences between parts of animal classes. 
% The alignment warps the source image into the target image.
% We compare Proposal Flow + \methodname (ours - 2nd row) and Proposal Flow + HoG \cite{ham2016} (3rd row).
% \rk{todo: add better legend}.
% }
% \label{fig:animalwarp}
% \end{figure*}

{\small
\bibliographystyle{ieee}
\bibliography{refsfixedCR}}

% --------------------------------------------------------------------

\clearpage

\small
\bibliographystyle{ieee}
\bibliography{refsfixedCR}

% --------------------------------------------------------------------
\end{document}